\documentclass[10pt,twocolumn,letterpaper]{article}

\usepackage[pagenumbers]{cvpr} 
\usepackage{times}
\usepackage{graphicx}
\usepackage{amsmath}
\usepackage{amssymb}
\usepackage{comment}
\usepackage{enumitem}
\usepackage[pagebackref]{hyperref}

\begin{document}

\title{Towards Billion-scale Multi-modal Biometric Search}

\author{Arka Koner\footnotemark[1] \quad Chetan S. Naik\footnotemark[1] \quad Lokesh Kurre\footnotemark[1] \quad Vivek Raghavan\footnotemark[1] \\
Barada P. Sabut\footnotemark[1] \quad Tanusree Deb Barma\footnotemark[1] \quad Anoop M. Namboodiri\footnotemark[2] \quad Anil K. Jain\footnotemark[3]}

\maketitle

\begin{abstract}
   Searching a multi-biometric database of a billion records for a country-level identity system requires pushing the limits of all aspects of a biometric system, including acquisition, preprocessing, feature extraction, accuracy, matching speed, presentation attack detection, and handling of special cases (\eg, missing finger digits). This is the first paper that gives insights into such a large-scale multi-modal biometric search system, called \emph{Bharat ABIS}, based on open-source architectures. The end-to-end pipeline of Bharat ABIS processes fingerprint, face and iris modalities through modality-specific stages of preprocessing (segmentation), quality assessment, presentation attack detection, and learning an embedding (feature extraction), producing a concatenated template of 13.5KB per person. We present a detailed analysis of the modalities and how they are integrated to create an efficient and effective solution for 1:N search (de-duplication). Evaluations on a demographically stratified gallery of 220 million identities, randomly sampled from 1.55 billion records in India's Aadhaar database, yield an FNIR of $\textbf{0.3\%}$ at an FPIR of $\textbf{0.5\%}$, for adult probes ($\geq18$ years). We also compare the performance of Bharat ABIS against three state-of-the-art COTS systems on a 20M gallery. Our system achieves a throughput of 100 searches per second on a gallery of 40M on a single server (8$\times$Nvidia H100 GPUs, 2TB RAM).
\end{abstract}


\footnotetext[1]{$^*$Unique Identification Authority of India, (\{arka.koner, chetan.naik, lokesh.techexe.tc, vivek.raghavan, hoe.dev-tc, tanusree.db\}@uidai.net.in)}
\footnotetext[2]{$^\dagger$IIIT Hyderabad, (anoop@iiit.ac.in)}
\footnotetext[3]{$^\ddagger$Michigan State University, (jain@cse.msu.edu)}


\section{Introduction}

\begin{figure}[t]
\begin{center}
   \includegraphics[width=\linewidth]{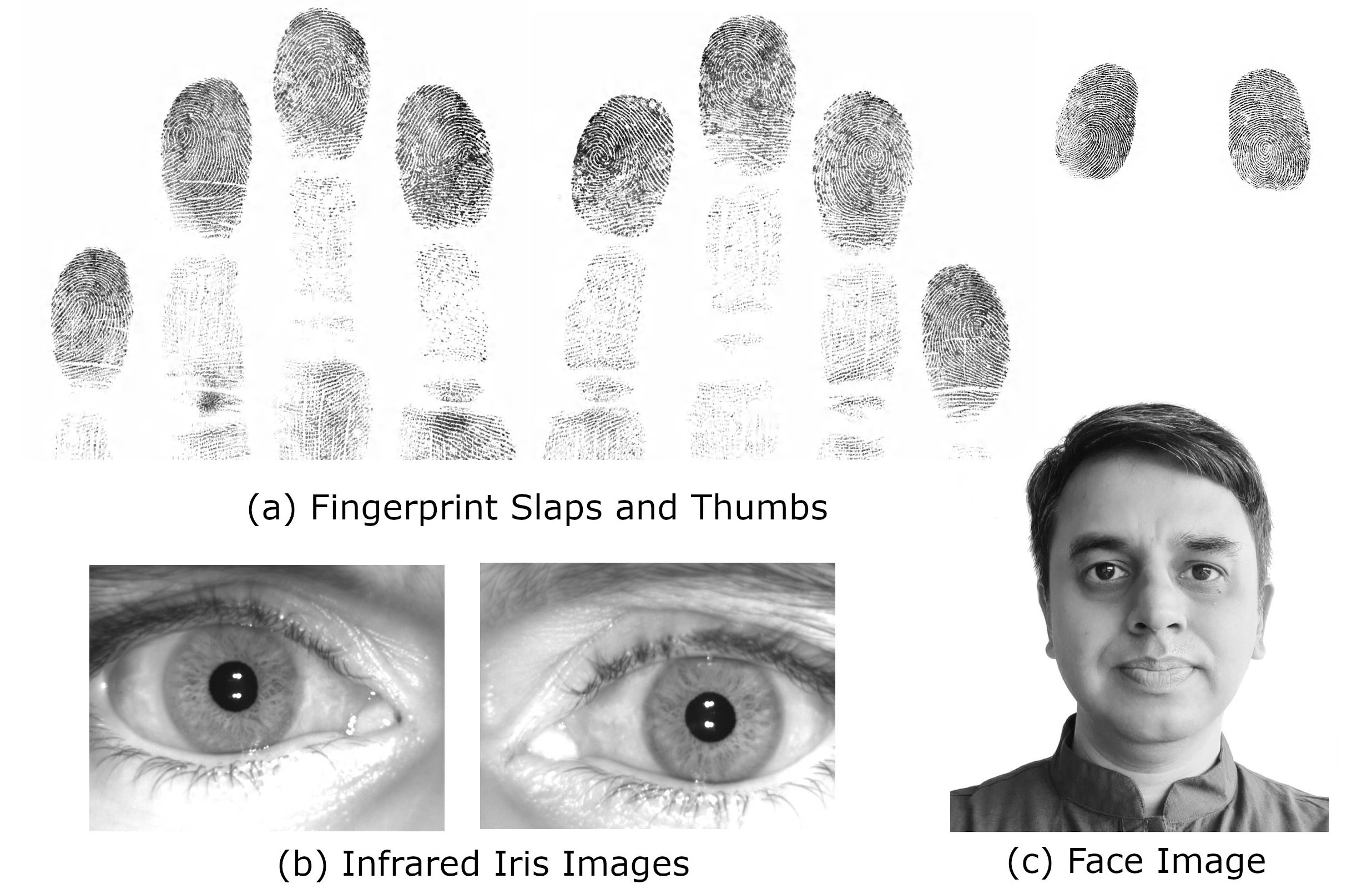}
\end{center}
   \caption{Biometric modalities used in Bharat ABIS: (a) 10-fingers (captured as 2 slaps and a thumb-pair), 2 irides and face. These modalities were selected to address the scale of the gallery (over 1.5 billion),  ease of capture, and search accuracy requirements (FPIR and FNIR).}
\label{fig:modalities}
\end{figure}

Identity documents such as passports, national identity cards (IDs), voter-IDs and driver licenses must ensure that a person never receives more than one identity, \ie, the ID number is, indeed, unique (one and only one ID number per person). The most reliable automated method to ensure uniqueness is to compare the biometric traits of every user in the database (gallery) against each other to determine if they belong to the same person or not. This process of identifying duplicates in a biometric database and retaining only one or linking them together is referred to as ``de-duplication''. In this paper, we look at the de-duplication problem in the context of a billion-scale national identity system, namely India's Aadhaar\footnote{UIDAI (Unique Identification Authority of India) is the statutory authority and government agency responsible for issuing Aadhaar, a 12-digit unique identity number: \url{https://uidai.gov.in/en/}}. The mandate of Aadhaar is to ensure that every resident in India is issued a verifiable 12-digit unique ID\footnote{In this article, we use the word ``ID'' to refer to the identity number that is linked with a user's biometric and demographic details; one or more physical or digital identity cards may be created for the same ID as long as the ID number remains the same.}, upon their enrollment. That is, no Indian resident should be issued a duplicate ID or is denied an ID irrespective of the ability of the system to capture their biometric traits or the availability of other identity documents such as birth certificate (Aadhaar Act~\cite{aadhaar-Act}).

\begin{figure}[ht]
\begin{center}
   \includegraphics[width=\linewidth]{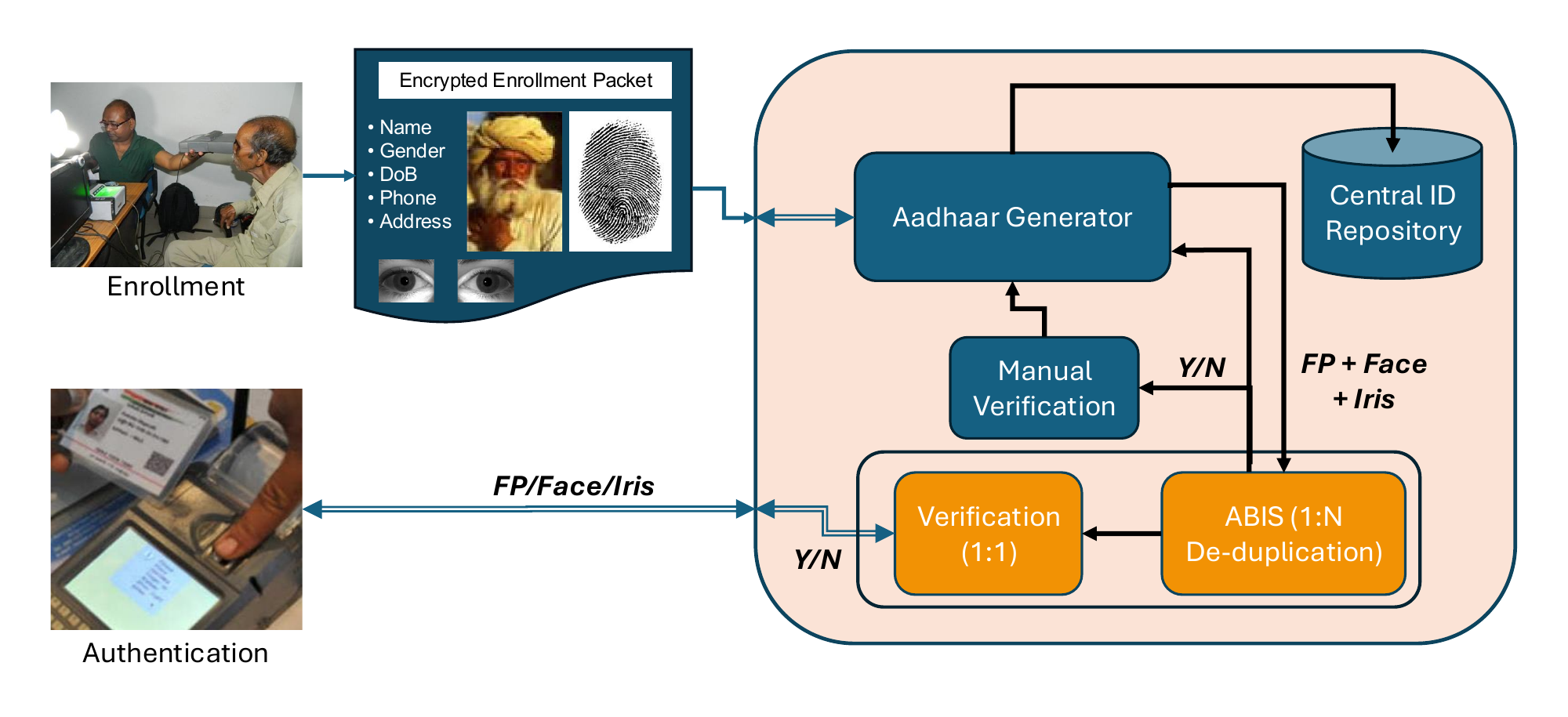}
\end{center}
   \caption{Data flow and interfaces of Aadhaar ABIS ( (Automated Biometric Identification System). The biometric data is encrypted at the capture device and remains encrypted in transit and at rest (storage) inside the Central ID Repository (CIDR). Once the biometric data enters the CIDR, it never leaves the system. The output of the ABIS module is a simple Yes/No decision during both enrollment (whether a duplicate is found in the gallery) and authentication (whether the query biometrics match the claimed identity).}
\label{fig:Aadhaar}
\end{figure}

The  Aadhaar Act\cite{aadhaar-Act} and the Digital Personal Data Protection (DPDP) Act~\cite{dpdp-Act} form the foundation on which the Aadhaar system is built. To ensure that deduplication is possible on a scale greater than 1.5 Billion, a pilot study was conducted in 2009 that evaluated the three most commonly used biometric modalities, namely face, fingerprint, and iris (see Figure~\ref{fig:modalities}). Recall that in 2009, the SOTA accuracy for face recognition was relatively low. So, it is not surprising that the study concluded that the use of all 10 fingerprints and 2 irides for all Indian residents is essential to ensure the desired de-duplication accuracy and population coverage. Although face was also captured for human verification, it was not used in the initial years for de-duplication due to its lower accuracy compared to fingerprint and iris. With the multi-fold increase in face recognition accuracy over time, the current Automated Biometric Identifications (ABIS) systems including Aadhaar use face images for both de-duplication and authentication. It is important to note that the Aadhaar act explicitly prevents the publication and sharing of any biometric data present in the Aadhaar database. While anonymized Aadhaar data was used for training and evaluation of all systems that we report\footnote{Models trained using public-domain databases were provided to UIDAI who then internally fine-tuned them  on subsets of privileged Aadhaar database.}, all images displayed in this paper are from other public sources or synthetically generated, and were chosen to illustrate data variability and failure cases. Figure~\ref{fig:Aadhaar} shows the primary functionalities provided by the Aadhaar system: enrollment, deduplication (1:N biometric search) and authentication  (1:1 comparison). The encrypted enrollment packet contains both biometric (face, fingerprints, and irides) and demographic (name, DoB, address, etc.) details of a resident.

The primary contributions of this work are as follows.
\begin{enumerate}
    \item A detailed analysis of design constraints in prototyping a billion-scale multimodal biometric identity system using open-source models.
    \item The design of a multi-modal ABIS, called Bharat ABIS\footnote{\textbf{Bharat} is an ancient name for India with deep historical, cultural, and linguistic roots.},  that is competitive to COTS (on a 20M gallery) in terms of accuracy and speed using open source DNN models for face, fingerprint and iris.
    \item Experimental evaluation of individual modalities and the multimodal system on a real-world multimodal biometric dataset of up to 220 million identities.
    \item Comparison of the proposed ABIS against three COTS systems currently operating in Aadhaar to establish State-of-the-art accuracies and search speeds.
\end{enumerate}

\subsection{Prior Work}

NIST periodically conducts biometric technology evaluations that define the state of the art in large-scale unimodal biometric identification: FpVTE for fingerprint~\cite{grother-FpVTE}, FRVT/FRTE for face~\cite{grother-FRVT}, and IREX for iris~\cite{IREX-9}. These evaluations benchmark individual modalities on galleries up to 16 million identities. However, NIST does not conduct multi-modal search (necessary for a national ID system), or the demographic diversity and scale of population present in a country like India. 

On the algorithmic side, several key advances underpin our system. DeepPrint, proposed by Engelsma \etal~\cite{engelsma-DP}, is a two-branch CNN for fingerprint representation that combines a texture branch for spatial features and a minutiae branch for domain knowledge, producing a 192-dimensional embedding that enables significantly faster inner-product matching compared to variable-length minutiae matching. For face recognition, FaceNet~\cite{schroff-FaceNet} is one of the simplest and most popular face representations. ArcFace~\cite{deng-ArcFace} introduces an additive angular margin penalty in the softmax classification loss, producing highly discriminative embeddings on a unit hypersphere. The InsightFace framework~\cite{an_2022_cvpr} provides a complete training pipeline around this loss function with standard backbones. For iris recognition, Daugman's phase-based IrisCode~\cite{daugman2004iris} is the classical method; recent deep embedding approaches with metric learning loss offer competitive accuracy with native FAISS compatibility.

For large-scale search, FAISS~\cite{johnson-BSS} provides a GPU-accelerated approximate nearest neighbor algorithm (flat index), which performs exact distance computation within Voronoi cells at sub-linear search complexity. This has been validated for a gallery of $10^9$ vectors. For fingerprint quality, NFIQ2~\cite{tabassi-NFIQ} is commonly adopted, although its correlation with match scores is algorithm-dependent. For fingerprint presentation attack detection (PAD), the Spoof Buster framework~\cite{chugh-SpoofBuster} demonstrated the effectiveness of minutiae-centered local patches. Ross and Jain~\cite{ross-Fusion} presented a taxonomy of multi-modal fusion strategies at sensor, feature, score, and decision levels.

While NIST-organized vendor evaluations have advanced unimodal biometric identification on galleries of up to tens of millions, no prior work reports a fully multi-modal de-duplication system designed and validated at a scale of hundreds of millions on a real-world national ID dataset with large demographic diversity.


\section{Design of Bharat ABIS}

A biometric identity system should be able to enroll a new user (1:N de-duplication) for large N and verify the claimed identity (1:1 comparison) of an enrolled user with high accuracy under varying biometric quality. The authentication process is expected to be light-weight with a near real-time response, while the template extraction may be done at the client-side. However, for de-duplication, an ABIS needs to employ a salient and robust feature extractor that maximizes search accuracy and search speed, even at the expense of template extraction speed. De-duplication and authentication often use different representations for a biometric modality, and one can extract the light-weight feature vector for authentication after the enrollment and de-duplication processes are completed.  While the ABIS might use multiple modalities for the de-duplication process, the authentication system might only use a single finger, iris or a face image. Most of the 100 million daily authentications in Aadhaar use a single biometric modality (primarily one fingerprint or face) along with the associated 12-digit Aadhaar. Among the world's top-4 most populous countries (India [1.47B], China [1.41B], United States [349M] and Indonesia [287M]), India's multimodal biometric national ID, called Aadhaar, successfully operating since 2009, has become a model for large-scale de-duplication and authentication.  In this paper, we focus on the design and evaluation of the proposed Bharat ABIS\footnote{While China has a national ID system that incorporates biometrics, we could not find its details. United States does not have a national ID (SSN is not issued based on biometrics). Indonesia has a 16-digit unique ID Number linked to fingerprints, iris, and face for citizens/residents 17 or older. \url{https://setkab.go.id/en/the-site-to-check-identity-card-isnt-created-by-ministry-of-home-affairs/}}.

The ABIS itself consists of three components: a) A biometric pipeline that processes each modality to create a template, b) a similarity search module that compares the template of a probe to all the identities in the gallery, and c) a scheduler that distributes and co-ordinates the tasks across different servers to handle 1 to billion de-duplication. In this work, we focus on the biometric pipeline. We have used the FAISS~\cite{douze-faiss} library for similarity search, which distributes the task across multiple GPUs~\cite{johnson-BSS} and has been customized for multi-node orchestration across servers.

\subsection{Design Constraints}

There are several real-world constraints that guided the design of the Aadhaar eco-system, which in turn dictated many of the design parameters of the biometric search solution. Below are some of the primary factors that resulted in our design decisions for Bharat ABIS.
\begin{enumerate}
    \item The Aadhaar Act mandates the issue of an identity to every Indian resident who enrolls. Moreover, death registrations are difficult to verify, so deceased individuals are not deleted from CIDR. These factors push the gallery size to over 1.5 billion.
    \item The above mandate also means that one has to enroll a person in spite of the quality of biometric acquisition or even the presence of all fingers and iris. In addition, a significant portion of residents and primary beneficiaries of a biometric ID system in a developing country like India are manual laborers with worn out fingerprints and cuts and bruises on the fingers. Hence, we cannot set a minimum quality threshold for fingerprints and reject an enrollment. This also requires the use of multiple modalities for de-duplication accuracy and population coverage.
    \item The enrollments and authentications are initiated from remote geographical locations in harsh weather conditions and limited internet connectivity. Hence, the capture devices used and the protocols employed for data transfer should work under these conditions.
    \item To ensure maximum population coverage, the cost of enrollment and authentication should be minimal. This required that enrollment has to happen through third-party agencies (referred to as registrars), and hence the process of data capture is not under the direct control of UIDAI. This requires operator training and setting standards for biometric capture~\cite{uid-bioStand}\cite{iso-bio}. In addition, biometric matching algorithms should be as efficient as possible and run on off-the-shelf hardware to keep infrastructure cost to a minimum.
    \item To enroll everyone in the country within a period of 3-5 years, at its peak, India's Aadhaar was enrolling over 1 million residents per day. This level of throughput was unprecedented. If a new ABIS is selected, it must first generate new templates from the billion-scale  enrolled data, not an easy task.
    \item As the biometric based national ID system forms the foundation for welfare disbursement, it is a target for hackers to create duplicate identities and fraudulent authentications via biometric spoofs. A presentation attack detection (PAD) and fraud detection system should be designed and continuously updated to handle evolving identity frauds.
\end{enumerate}

\subsection{Biometric Acquisition}

As discussed, each Aadhaar enrollment captures three biometric modalities: (i) fingerprints as three slap impressions (left four fingers, right four fingers, two thumbs) on FTIR slap scanners at 500 dpi; (ii) left and right iris images captured simultaneously in near-infrared; and (iii) a frontal face photograph in the visible spectrum. Each modality is acquired multiple times to ensure the highest quality biometric data. This yields 10 fingerprint images, one per digit, 2 iris images, and 1 face image per resident.

\subsection{Biometric Data Processing}

\begin{figure}
\begin{center}
   \includegraphics[width=\linewidth]{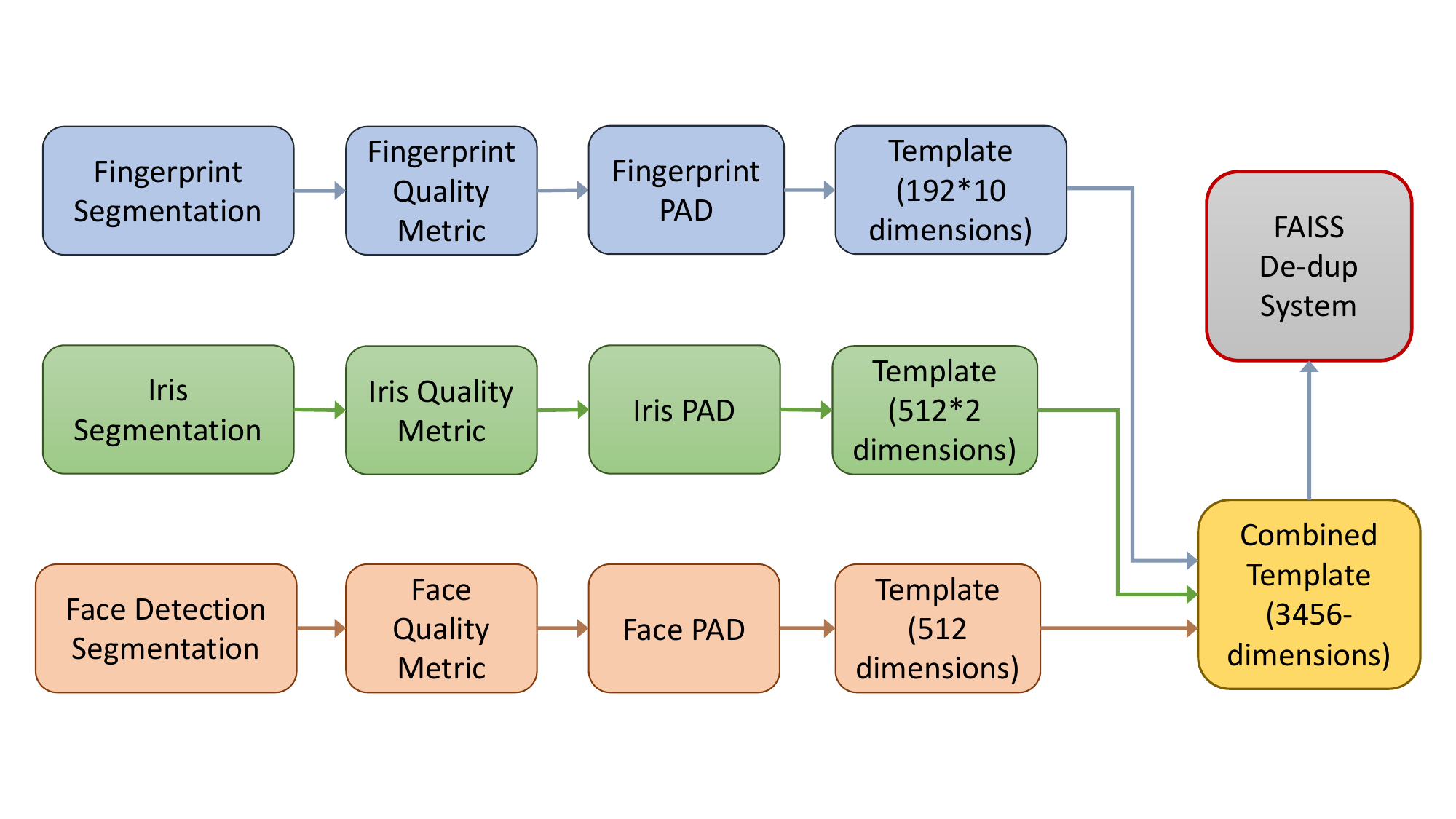}
\end{center}
   \caption{Multi-modal template generation pipeline. Each modality passes through segmentation, quality assessment and PAD, before embedding extraction. The embeddings of all modalities are concatenated into a single template vector of 3,456 dimensions (13.5KB). Comparison of a probe template against the gallery is done using the Facebook AI Similarity Search (FAISS)~\cite{douze-faiss} system.}
\label{fig:ABIS-Pipeline}
\end{figure}

We now describe the design of the biometric processing modules in Bharat ABIS in detail. The architecture follows a sequential data processing design (see Figure~\ref{fig:ABIS-Pipeline}) with exception handling in which a query identity is converted to a fixed length representation before the search for a duplicate in the gallery is performed.

The first three stages of fingerprint, face and iris modalities are image segmentation/detection, biometric quality estimation and presentation attack detection. The data capture station will mark any exceptions in capture of the biometrics such as failure to acquire, which is checked first. The biometric impression is localized within the captured image.
 All deep learning models described below are image-based open-source convolutional architectures trained on anonymized UIDAI enrollment data. The training datasets for segmentation and quality assessment had approximately 200K unique subjects with around 300K samples per modality, drawn to represent the demographic diversity of the Indian population across age, gender, geographical location, ethnicity, and acquisition device variability.

\subsubsection{Image Segmentation}

For face images, a detector based on YOLOv7~\cite{wang-Yolov7} was developed based on the image characteristics captured  for Aadhaar. The detector localizes the face bounding box and extracts five facial landmarks (eye centers, nose tip, mouth corners). An affine transformation normalizes the detected face to a canonical 112 × 112 × 3 input, compensating for pose and alignment variations. 

For fingerprint images, there are multiple bounding boxes for each slap and thumb images. The same YOLOv7 architecture is used to create a fingerprint-specific segmentation module that detects all distal phalanges, trained on a dataset of around 100K annotated slap images. The proposed segmentation scheme shows improved robustness on degraded-quality captures. The same segmentation output is used for matching during authentication requests.

For Iris segmentation, we employ an EfficientNet-based~\cite{tan-Efficientnet} U-Net architecture that generates pixel-level segmentation masks identifying the pupillary boundary, limbic boundary, and eyelid/eyelash occlusions. Hough gradient-based circle fitting is applied to the predicted mask to estimate precise circular parameters for the inner (pupil) and outer (limbus) boundaries.

\subsubsection{Image Quality Estimation}

A poor quality fingerprint has limited matching utility: authentication against such a sample is likely to fail regardless of the matcher. We use UIDAI Fingerprint Image Quality (UFIQ), a quality predictor built on NFIQ2~\cite{tabassi-NFIQ} but retrained as a Random Forest regressor specifically for FTIR slap scanners used in the Aadhaar ecosystem. UFIQ was trained to predict the authentication match score. It achieves a Pearson correlation of $r = 0.78$, compared to $r = 0.54$ for standard off-the-shelf NFIQ2. This $44\%$ relative improvement in predictive power enables quality-adaptive routing within the pipeline: high-quality samples proceed directly to embedding extraction, while low-quality samples may be flagged for enhancement or re-acquisition at the enrollment station.

For face and iris images, custom CNN-based quality estimators are trained that evaluate several quality parameters. The face quality module  evaluates normalized face images for factors such as illumination, motion blur, noise, resolution, and pose. The iris quality module evaluates factors such as degree of occlusion by eyelids or eye-lashes, pupil dilation ratio, off-angle gaze, motion blur, and image sharpness. As there is no universally accepted quality standard for face and iris (unlike NFIQ for fingerprint), these modules were developed and calibrated internally using correlation with operational matching scores. Quality scores determine downstream fusion weighting: high-quality samples receive greater weight in the score integration stage.

\subsubsection{Presentation Attack Detection}

Although the Presentation Attack Detection (PAD) algorithms do not directly influence the de-duplication process, it plays an important role in preventing fraudulent enrollments. Specifically, fraudulent attempts such as toe prints, slap prints with middle phalanges, or mixed fingerprints should be detected. Common spoof attempts include — for face: replay attacks and eye-cut photos; for iris: artificial eyeballs, printed paper, and contact lenses. Such attempts can be used to enroll a new identity in order to fraudulently obtain several benefits, such as from social service programs. Since such atypical or mixed biometric inputs are generally absent from the gallery, the de-duplication process may fail to identify them as duplicates, thereby allowing a successful — but fraudulent — enrollment.

The fingerprint PAD module verifies that each biometric sample originates from a live finger. The liveness classifier follows the minutiae-centered local patch approach of the Fingerprint Spoof Buster framework~\cite{chugh-SpoofBuster}: a MobileNetV1 model is trained on small $96{\times}96$ patches extracted around detected minutiae points. Training data includes a balanced in-house dataset (30K live, 30K non-live samples), the LivDet 2015 benchmark~\cite{yambay-LivDet}, and labeled production data. Non-live fabrication materials cover silicone, polymer, latex, wax, wood glue, Fevicol, and other substrates encountered operationally. The fingerprint PAD model achieves a True Detection Rate (TDR) of $99.5\%$ (when we correctly flag a spoof input) at a False Detection Rate (FDR) of $0.5\%$ (when a live input is declared as spoof) on field-tested ECMP slap scanner images. A separate CNN-based non-distal phalange detector identifies cases where intermediate phalanges are presented instead of fingertips. Additional modules detect slap swapping, toe-print enrollment, and mixed-biometric packets.

For faces, a MobileNetV1-based liveness module distinguishes genuine face presentations from spoofing attacks across four categories: print attacks (printed photographs), eye-cut photo attacks (printed photo with eye regions cut out and real eyes placed behind), wrapped photo attacks (photo bent to simulate 3D curvature), and video replay attacks. The model achieves a TDR of $99.5\%$ at an FDR of $0.5\%$ on the Aadhaar evaluation set.

The iris liveness classifier detects presentation attacks that include printed iris images, patterned contact lenses, prosthetic eyes, and screen replay. This CNN-based model was trained on both public spoofing databases and Aadhaar's operational fraud cases. The module achieves a TDR of $99.0\%$ at an FDR of $1.0\%$ on internal evaluation.

\subsection{Fixed-Length Embedding Generation}
This stage of ABIS transforms the processed biometric images into fixed-length feature vectors (embeddings). 
\subsubsection{Fingerprint}
The core fingerprint representation is produced by DeepPrint~\cite{engelsma-DP}, a two-branch convolutional network. Both branches share a spatial transformer network (STN) that aligns the input fingerprint to a canonical pose. The texture branch then applies a base CNN to extract spatial features from the aligned image. The minutiae branch injects domain knowledge by adding a learned minutiae map to the feature representation. Each branch produces a 96-dimensional vector; their concatenation yields the final 192-dimensional per-finger embedding. The matching score between two fingerprint templates is the inner product of the two vectors, a computation executing in sub-millisecond time and natively compatible with FAISS-based nearest neighbor search. The DeepPrint model was trained from scratch on approximately 3M anonymized Aadhaar fingerprint images from over 200K unique subjects. Data augmentation comprised geometric transformations (rotations of $\pm 10$--$15^{\circ}$, translation, and scale jitter), random Gaussian blur with a small kernel, and random partial occlusion and cropping. The network was optimized with Adam (learning rate $10^{-4}$ to $10^{-3}$ under a step-decay schedule) using a composite loss combining center loss, cross-entropy loss, and mean squared error (MSE). Training was performed on NVIDIA V100 GPUs. Single-finger verification accuracy is $97\%$ TMR at $FMR = 0.01\%$. When we use all 10 fingers (concatenated embedding size 1,920), the system achieves $TMR = 96.25\%$ at $FMR = 10^{-9}$, the extreme operating point required for billion-scale 1:N search where even a very small per-comparison false match rate produces an unacceptable absolute number of false candidates.
\subsubsection{Face}
Face embeddings are extracted using a ResNet-50 backbone, pretrained on MS-Celeb~\cite{guo2016msceleb}, and trained with ArcFace additive angular margin loss~\cite{deng-ArcFace} (scale $s = 64$, margin $m = 0.5$) following the InsightFace framework~\cite{an_2022_cvpr}. ArcFace modifies the standard softmax classification loss by introducing an angular margin, \emph{m}, between the feature vector and its corresponding class weight on the unit hypersphere, enforcing tighter intra-class compactness and wider inter-class separation. The trained model produces a 512-dimensional float embedding per face image. Matching between two face templates is performed as the inner product of the two embedding vectors. The backbone was fine-tuned on 300K anonymized Aadhaar face images from 150K subjects, with warp-affine augmentation following InsightFace-style landmark-based alignment. Optimization used the Adam optimizer (learning rate $10^{-4}$ to $10^{-3}$ under a step-decay schedule) on NVIDIA V100 GPUs. The model achieves $TMR = 99.5\%$ at $FMR = 0.01\%$ on internal evaluation data.
\subsubsection{Iris}
Iris embeddings are extracted using a ResNet-50 backbone, initialized from ImageNet pretraining, and trained with ArcFace loss (scale $s = 64$, margin $m = 0.5$) following the InsightFace framework, producing a 512-dimensional embedding per iris image. The model was trained on approximately 3M anonymized UIDAI iris images from over 200K subjects, with augmentation comprising small rotations, brightness/contrast adjustments, and blur. Optimization used the Adam optimizer (learning rate $10^{-4}$ to $10^{-3}$ under a step-decay schedule) on NVIDIA V100 GPUs. Matching is via inner product. The system achieves $TMR=96.85\%$ at $FMR=0.01\%$ on internal evaluation data. The model parameters for each modality were determined empirically, by evaluating performance of the trained model on an independent validation set.

\subsection{Multi-Biometric Template}

The multi-biometric template is a single concatenated template per identity:
$t = [f_1; \ldots; f_{10}; v; r_L; r_R] \in \mathbb{R}^{3456}$
where fi $\in$ $\mathbb{R}^{192}$ is the DeepPrint embedding for the i-th finger, v $\in$ $\mathbb{R}^{512}$ is the face embedding, and rL, rR $\in$ $\mathbb{R}^{512}$ are the left and right iris embeddings. The dimensionality decomposes as $10 \times 192 + 512 + 2 \times 512 = 3,456$, which results in a 13.5KB feature vector (4 bytes per dimension).

\subsection{Gallery Search and Score Fusion}
After template generation, the 3,456-dimensional query template is submitted to the search engine for 1:N de-duplication. The inner product of the query template is computed against gallery templates using FAISS~\cite{douze-faiss} with IndexFlat indexing on GPU. The gallery is partitioned into shards of optimal size across distributed GPU servers for parallel search; per-shard results are aggregated.
The final matching score is computed as a weighted linear combination of per-modality inner products: $S=\sum_{mod} w_{mod} \cdot \langle v_q, v_g \rangle$. The weights for the modalities were in the ratio of Face~(12.5) : Iris~(6.25) : Thumb and Index Fingers~(2.3) : Other Fingers~(1). The weights were empirically determined on a validation dataset and reflect the relative discriminative power of each modality as well as the total length of features from each modality. The system supports age-specific and quality-adaptive weight configurations. A threshold-based decision on the fused score classifies each query as duplicate or unique. Candidates exceeding the duplicate threshold (around $0.1\%$ in practice) are surfaced for manual adjudication.


\section{Experimental Evaluation and Analysis}

The de-duplication evaluations were conducted on an anonymized gallery of $N = 20M$ identities. This 20M gallery was randomly sampled from $1.55$ billion records in the Aadhaar database with stratification to ensure representativeness in age, sex, ethnicity, and geographic region. The same data set is used as the standard benchmark to evaluate all three COTS systems currently utilized by Aadhaar and internal ABIS solutions, enabling direct comparability of results across the solutions. The gallery size is limited to 20M as COTS systems were not available for evaluation on larger datasets in benchmarking environment. Although the gallery contains all age groups ($\geq 5$ years), the query evaluation for this study focused on adult population (age $\geq 18$), for which the current models are optimized. The query set comprises $37,835$ known duplicate packets (manually verified) and $34,812$ known non-duplicate packets. The number of queries was chosen to ensure statistical significance of the results at the required accuracies of FPIR$\leq0.5\%$ and FNIR$\leq0.1\%$. 

\subsection{Dataset Challenges}
\begin{figure}[htb]
\begin{center}
   \includegraphics[width=\linewidth]{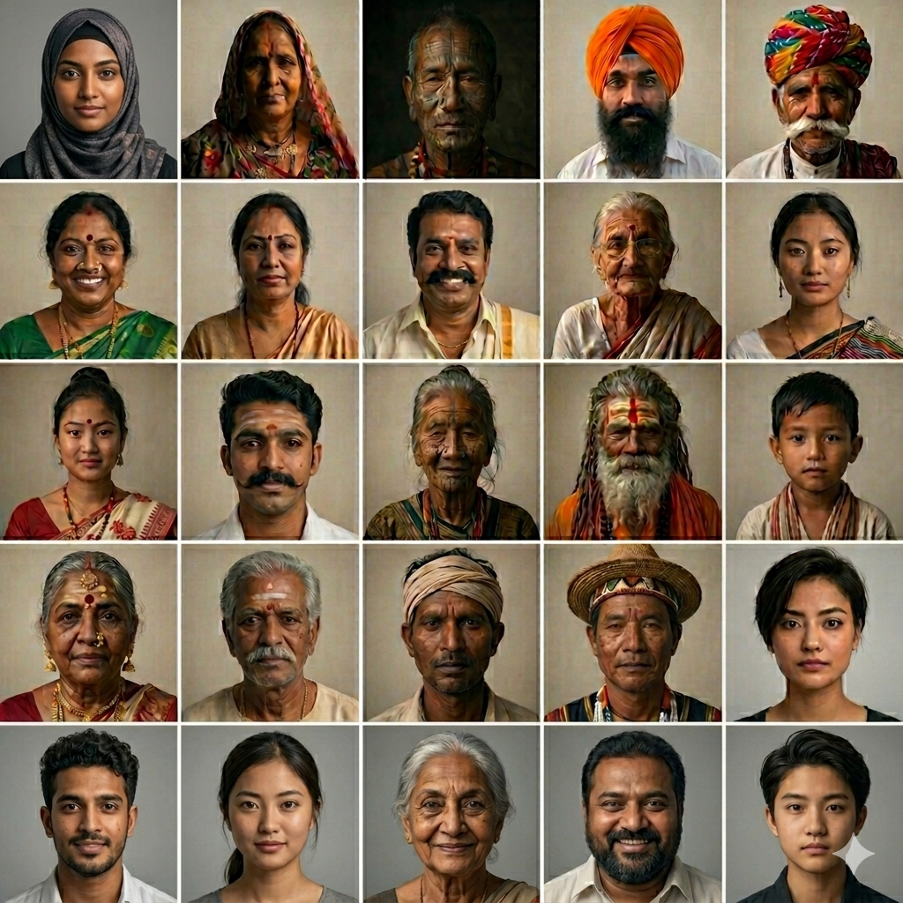}
\end{center}
   \caption{Face image variations due to ethnicity, social norms and religious practices. These are synthetic images that closely mimic the variability we observe in the Aadhaar dataset. The presence of head coverings, accessories and marks on the faces make the problem of face matching more challenging.}
\label{fig:faceData}
\end{figure}

As mentioned in the introduction, we are unable to provide image samples from the Aadhaar dataset due to privacy and legal considerations. To understand the challenges, Figure~\ref{fig:faceData} shows a set of synthetic face images we created to resemble a selection of face images from the Aadhaar dataset. Challenges in faces matching arise due to poor exposure, head coverings, accessories, ``bindis''  (dots on women's forehead), ``vihbooti'' (ash markings) on the forehead, etc.

\begin{figure}[htb]
\begin{center}
   \includegraphics[width=\linewidth]{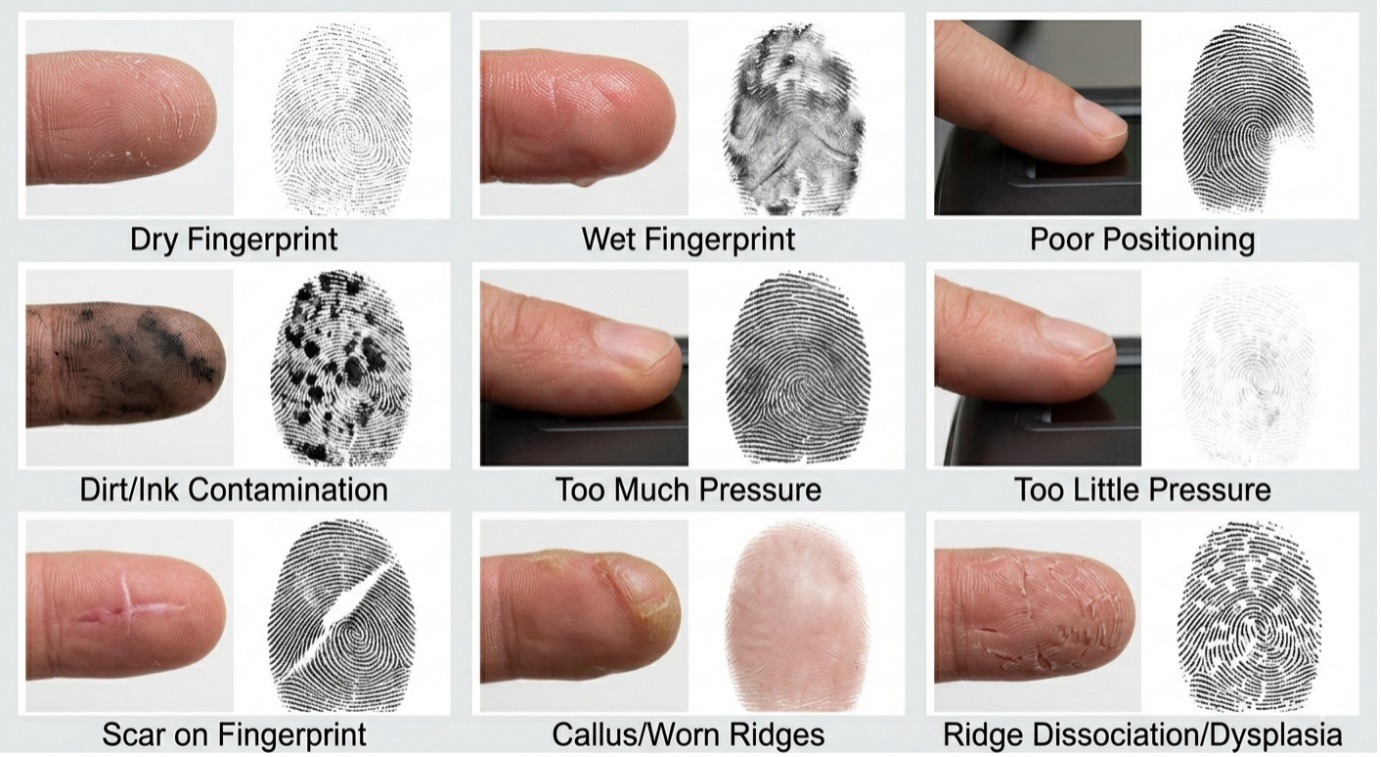}
\end{center}
   \caption{Examples of poor quality fingerprint images. These are synthetic images that closely resemble the challenges we observe in the Aadhaar dataset.}
\label{fig:fingerData}
\end{figure}

\begin{figure}[htb]
\begin{center}
   \includegraphics[width=\linewidth]{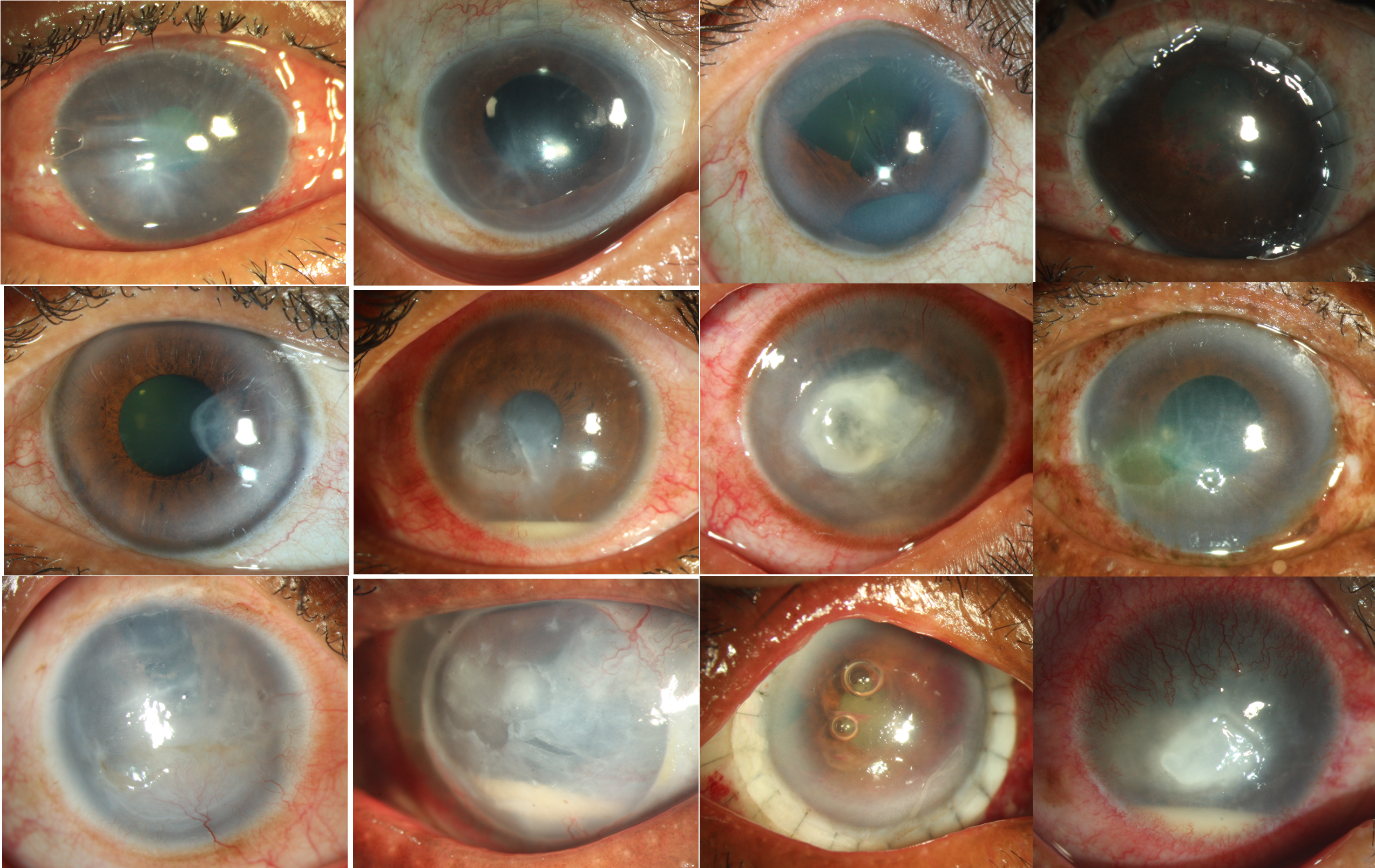}
\end{center}
   \caption{Examples of  commonly seen eye conditions that lead to poor quality Iris capture.}
\label{fig:irisData}
\end{figure}

For fingerprints, image quality suffers mostly from skin and capture conditions. Figure~\ref{fig:fingerData} shows a set of fingerprint images that closely resembles the poor quality data that we observe in the Aadhaar dataset.

The quality of iris capture suffers from several commonly seen eye conditions, including Glaucoma, Keratitis, Corneal Ulcer, etc. Figure~\ref{fig:irisData} shows examples of eye images that cause poor quality iris capture.

\begin{figure}[!htb]
\begin{center}
   \includegraphics[width=0.7\linewidth, height=1.8in]{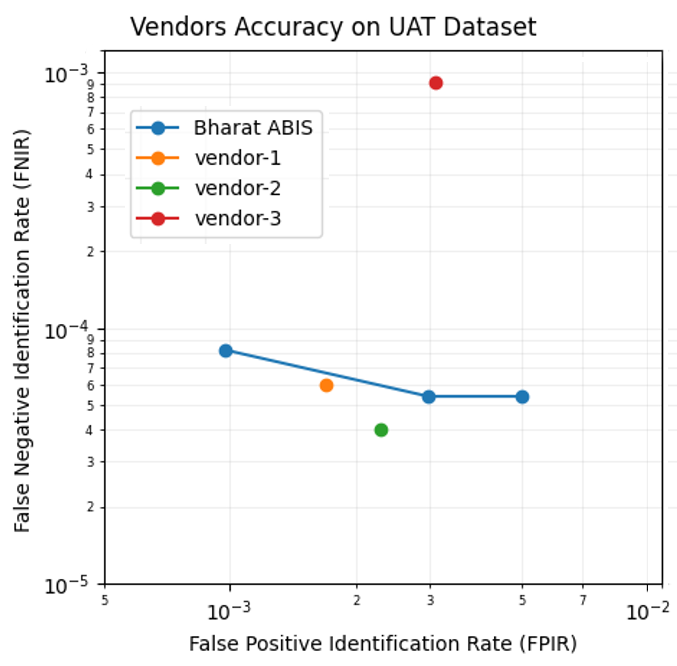}
\end{center}
   \caption{FPIR vs FNIR plot (log scale) of Bharat ABIS and three state-of-the-art COTS solutions who provided their performance at a single operating point. The gallery size is 20M subjects consisting of 10 fingerprints, face and 2 irides of every subject. Probe set consists of $\sim35K$ mated pairs and $\sim38K$ non-mated pairs.}
\label{fig:vendorComparison}
\end{figure}

Note that the evaluation of the FPIR-FNIR values of a real-world biometric de-duplication system that is in operation is challenging as we do not know the ground-truth (identity) of subjects that arrive for enrollment during regular operations. To overcome this, we create a set of mated ($\sim35K$) and non-mated ($\sim38K$) probes that are manually verified. This dataset along with an authentication test-set of 150K identities (referred to as the UAT dataset) is also used for comparison of COTS systems with our Bharat ABIS on the 20M gallery. The internal evaluation of COTS systems is done at a single operating point, and each system is tuned to achieve $FPIR \leq 0.5\%$ at $FNIR \leq 0.1\%$. Figure~\ref{fig:vendorComparison} shows that the proposed Bharat ABIS achieves comparable or better accuracy w.r.t. to the three COTS deployed by Aadhaar on the 20M gallery sampled from essentially, the whole population of India.

\begin{table}[ht]
    \centering
    \begin{tabular}{|c|c|c|c|c|}
    \hline
    Metric & \# Queries & \# Errors & Error Rate & Target \\
    \hline
    FPIR & $34,812$ & $35$ & $0.1\%$ & $\leq 0.5\%$ \\
    \hline
    FNIR & $37,835$ & $18$ & $0.05\%$ & $\leq 0.1\%$ \\
    \hline
    \end{tabular}
    \vspace{0.2cm}
    \caption{The number of mated (in all three modalities) and non-mated queries used for testing  COTS and Bharat ABIS systems on a 20M gallery. The observed FPIR and FNIR for Bharat ABIS are $5\times$ and $2.5\times$ better than the target rates for 20M gallery.}
    \label{tab:acc-Overall}
\end{table}

\subsection{Benefits of Multi-modality}

Table 1 shows that Bharat ABIS produces $35$ false positives out of $34,812$ non-mated queries and $18$ false negatives out of $37,835$ mated queries. Note that the factors that make the individual biometric modalities challenging also affect the overall accuracy of a multi-modal biometric matcher. In most cases, the qualities of individual fingerprints of a person are highly correlated and so are the conditions like cataract that affect both the eyes. The proposed Bharat ABIS achieves an FNIR of $4.76\times 10^{-4}$ @ FPIR$=0.1\%$, which implies that a fraudulent re-enrollment attempt is detected with probability $99.95\%$. Both metrics exceed the procurement specification set by Aadhaar for COTS solutions.

\begin{table}[ht]
    \centering
    \begin{tabular}{|c|c|}
    \hline
    Modality  & TMR @ FMR $=0.01\%$ \\
    \hline
    Face  &  $99.5\%$  \\
    \hline
    Iris  &  $96.85\%$ \\
    \hline
    Fingerprint &  $97\%$ \\
    \hline
    \end{tabular}
    \vspace{0.2cm}
    \caption{True Match Rate (TMR) for authentication using individual modalities (single finger, single iris, and face) at a False Match Rate (FMR) of $0.01\%$. Verification experiments were done on a dataset of 100K identities with 2 samples per identity.}
    \label{tab:accModalities}
\end{table}

\begin{figure}[ht]
\begin{center}
\includegraphics[width=\linewidth, height=2.0in]{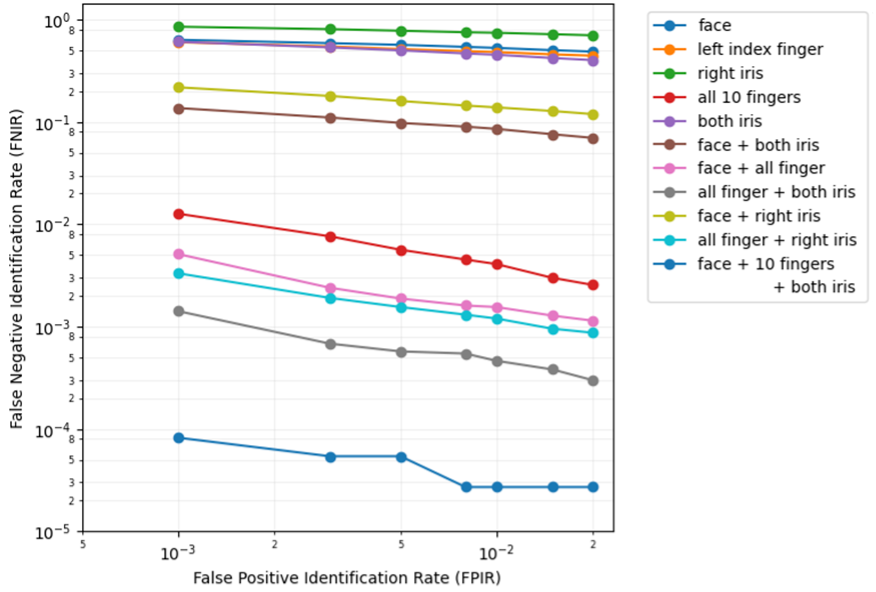}
\end{center}
   \caption{FPIR vs FNIR plots of individual modalities and their combinations (both axes on log scale). At FPIR of $0.1\%$, unimodal FNIR values are 63.6\% (Face), 60.3\% (Fingerprint), and 85.5\% (Iris). Combining all 10 fingers reduces FNIR to 1.3\%; two-modality combinations yield 13.7\% (Face+Irides), 0.51\% (Fingers+Face), and 0.14\% (Fingers+Irides). Fusion of all three modalities achieves FNIR of 0.01\%. There is more than an order of magnitude reduction in error rate from single modality to two-modality combinations, with further reduction when fusing all three-modalities. Gallery size is 20M with $\sim$35K mated and $\sim$38K non-mated samples.}
\label{fig:modalitiesErr}
\end{figure}

The accuracy of individual modalities and the improvements due to their fusion are two of the key aspects of any multi-modal de-duplication system. Let us take a closer look at the individual modalities in Figure~\ref{fig:modalitiesErr}. We notice that face and single finger (left index) are essentially equal in discrimination power, while single iris is a close second (note that we are showing FNIR vs. FPIR, so lower the curve, better the performance). Combining all 10 fingers gives a significant reduction is FNIR. Similarly, Combining two modalities at a time and all-three modalities offer a significant boost in performance.

\subsection{Effect of Gallery Size}

\begin{figure}[t]
\begin{center}
   \includegraphics[width=0.8\linewidth, height=2.0in]{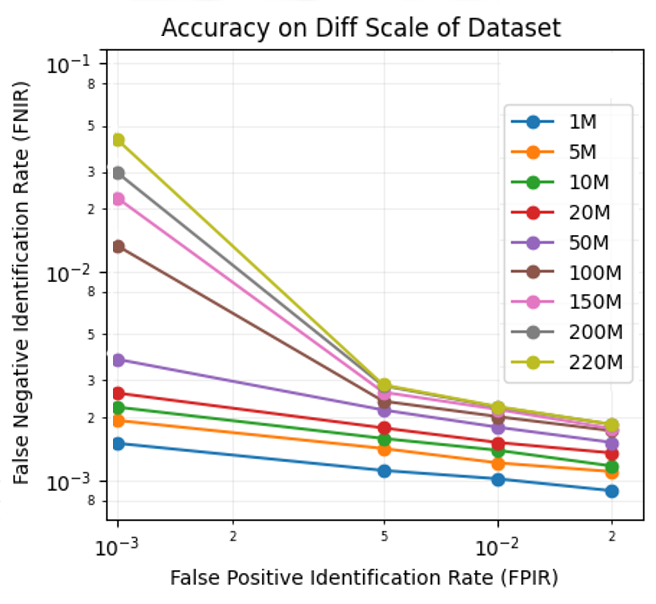}
\end{center}
   \caption{FNIR vs. FPIR plots for Bharat ABIS for various gallery sizes. At an FPIR of $1$ in $1000$, the FNIR (@gallery size) values are: 0.0015 (@1M), 0.002 (@5M), 0.0025  (@10M), 0.0028  (@20M), 0.004 (@50M), 0.012 (@100M), 0.022 (@150M), 0.03 (@200M), and .04 (@220M). As expected, for a fixed FPIR, FNIR increases with increase in gallery size. Both axes are on log scale.}
\label{fig:gallerySize}
\end{figure}

We know that FPIR increases with increasing gallery size as a larger number of non-mated comparisons raises the probability that at least one of these comparisons exceeds the decision threshold. To understand this effect in practice, we performed deduplication experiments on different gallery sizes, ranging from 1M to 220M. Figure~\ref{fig:gallerySize} shows an FNIR vs. FPIR plots for the different gallery sizes. We report these results only for the proposed Bharat ABIS. Carrying out these experiments on COTS is not possible because we cannot access them. 

As the gallery scales further from 200M to the full 1.5 billion identities in Aadhaar, the FPIR is expected to increase. The FNIR remains approximately constant at a given threshold (since it depends only on the quality of the mated comparison, and not on the gallery size). 

\subsection{Effect of Gallery Size}

\begin{figure}
\begin{center}
   \includegraphics[width=0.9\linewidth]{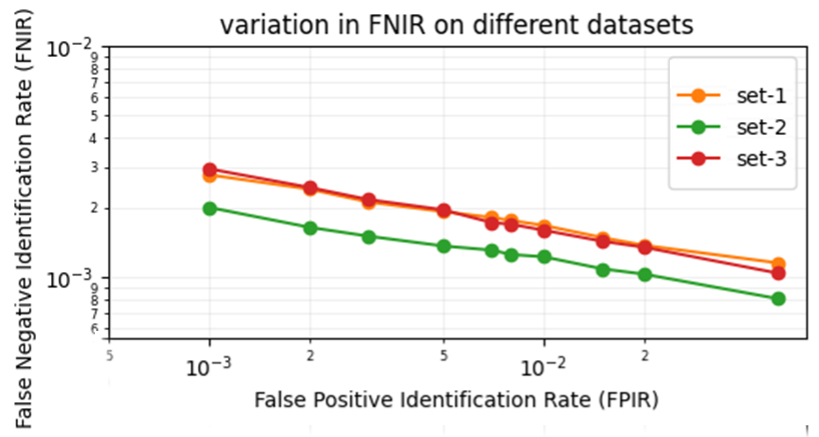}
\end{center}
   \caption{Effect of different 20M galleries randomly sampled from the 220M gallery from the Aadhaar database on accuracy. We note that three different subsets of 20M show only a small variation in performance.}
\label{fig:datasetVar}
\end{figure}

To understand the effect of gallery selection in these experiments, we repeat the selection of 20M identities from a larger dataset of 220M from 11 Indian states. Note that this 220M is a subset of the total population of India. The results, presented in Figure~\ref{fig:datasetVar} shows that the solution is robust to variations in the test set that are representative of the population.

\subsection{Search Speed and Throughput}
At a billion scale, search latency matters as much as accuracy. Each $3456$-dimensional template (13.5~KB) is held in GPU memory as an $N \times 3456$ single-precision matrix, and matching reduces to a batched inner product with top-$k$ selection -- FAISS flat-index (exact) search on GPU~\cite{johnson-BSS,douze-faiss}. On a single DGX server with 8$\times$NVIDIA H100 GPUs (80~GB each), 5M templates fit per GPU, giving 40M residents per server. Larger galleries are handled by streaming additional shards and aggregating top-$k$ results on the host. 

The search takes around 10~seconds per $1000$ probes against the 40M gallery, resulting in a throughput of 100 probes per second for exact 1:N search. Throughput scales near-linearly with batch size because the kernel is memory-bandwidth bound and each probe reuses the gallery matrix already in VRAM. Beyond 40M templates per server, shards run in parallel across servers and end-to-end latency is set by the slowest shard. Combined with score fusion described in Section~2.6, our Bharat ABIS meets the $\geq$1M de-duplications per day target with exact matching at the final stage.

Beyond de-duplication, DeepPrint is also deployed for Aadhaar fingerprint authentication, where fusion roughly halves the failure rate versus a vendor SDK alone. On a base of over 50 million daily authentications, this yields millions of additional successful transactions per day, with the largest gains for elderly users and manual labourers whose fingerprint quality is typically poor.

\subsection{Limitations and Mitigation Strategies}
The current models in Bharat ABIS are optimized for the adult population (18 years and older), while the production system is expected to handle all ages above 5. Face matching accuracy degrades for subjects below age 15
when the enrollment-to-query gap exceeds five years, reflecting the biological reality of
rapid facial development in children. Mitigation strategies include age-adaptive fusion
weights, GAN-based face age progression for data augmentation, and synthetic fingerprint
generation via diffusion models to expand training data diversity across sensor types.

\subsection{Lessons Learned}

The prototyping of Bharat ABIS has highlighted several practical aspects of building a large-scale multi-biometric ABIS. These include:
\begin{itemize}[leftmargin=*]
    \item Open source models can be used to build a large scale multi-modal biometric system.
    \item Face, fingerprint and iris are now the biometrics of choice; their fusion enables billion-scale ABIS as demonstrated in the success of Aadhaar since 2009. 
    \item Building highly accurate and large scale biometric systems to support social good, national ID and homeland security is the next frontier for biometric researchers.
    \item Fusion of open source architectures for face, fingerprint and iris recognition can provide close to SOTA accuracy for upto 100M identities. However, researchers must work collaboratively with user agencies (NGOs and government agencies) with access to large-scale biometric data so that these models can be either trained from scratch or finetuned on large domain-specific datasets. The workflow we followed to achieve high accuracy without compromising data privacy is given in the supplementary material.
    \item Our training sets for face, fingerprint, and iris for Bharat ABIS were of the order of 100K unique identities compared to a test set gallery size of 220M. 
    \item Integrity, quality, and standards for data acquisition both at enrollment and recognition are key to success; and
    \item Given that over 100 nations (more than half of UN-recognized states) have populations under 10 million, the proposed Bharat ABIS  model can be deployed in these countries without any difficulty. Many of these countries cannot afford a COTS ABIS.
\end{itemize}

\section{Conclusions and Future Work}

The proposed Bharat ABIS effectively integrates real-world constraints with embeddings (features) learned from representative large-scale biometric datasets. The system uses 10-fingers, 2-irides and face modalities to achieve an FPIR of $\textbf{0.10\%}$ at an FNIR of $\textbf{0.05\%}$ on a gallery of 20 million demographically representative identities, exceeding the Aadhaar design requirements of FPIR$\leq 0.5\%$ at FNIR$\leq 0.1\%$. World Population Review~\cite{wpr26} estimates show that there are over 160 countries and independent territories in the world with a population of fewer than 20 million people. Hence, we believe Bharat ABIS can be easily deployed in these countries, many of which cannot afford to purchase a COTS ABIS. The proposed fingerprint templating solution, deployed for authentication, has also halved authentication failure rates in the field. Our research and prototyping of Bharat ABIS demonstrates that one can build a sovereign, high-performance, multi-modal biometric de-duplication system on open-source AI foundations. Scaling the solution to the entire gallery of 1.55 billion is underway, but will require significant innovations in model design, training and engineering.

\subsection{Acknowledgements}

We acknowledge the open-source communities behind TensorFlow, PyTorch, FAISS, YOLOv7, and InsightFace. We also thank Akshat Singhal, Mahizhvannan E., Kritika Gupta, Deepak Kumar Sial, and Sharib Athar for their valuable contributions.

{\small
\bibliographystyle{ieee}
\bibliography{egbib}
}

\end{document}